\documentclass[letterpaper, 10 pt, conference]{ieeeconf}  

\IEEEoverridecommandlockouts          
\usepackage[pdftex]{graphicx}
\graphicspath{{../pdf/}{../jpeg/}}
\DeclareGraphicsExtensions{.pdf,.jpeg,.png}

\usepackage{algorithm}
\usepackage{amsmath}
\usepackage[noend]{algpseudocode}
\usepackage{euscript}
\usepackage{color}
\usepackage{booktabs} 

\makeatletter
\let\NAT@parse\undefined
\makeatother
\usepackage[numbers,sort,compress]{natbib}
\usepackage{amsfonts}
\usepackage{mathtools}
\usepackage{placeins}
\usepackage{microtype}
\usepackage{soul}
\usepackage{tikz}

\usepackage[caption=false,font=footnotesize]{subfig}
\usepackage[hidelinks]{hyperref}
\usepackage[capitalise]{cleveref}
\usepackage{fancyhdr}

\crefformat{section}{#2Sec.~{\textup{#1}}#3}
\crefformat{figure}{#2Fig.~{\textup{#1}}#3}
\crefformat{equation}{#2({\textup{#1}})#3}
\crefmultiformat{figure}{Figs.~#2#1#3}{ and~#2#1#3}{}{}
\crefmultiformat{section}{Secs.~#2#1#3}{ and~#2#1#3}{}{}
\crefrangeformat{section}{Secs.~#3#1#4 to~#5#2#6}

\newcommand{\norm}[1]{\left\lVert#1\right\rVert}

\newcommand{\graph}{\mathcal{N}}
\newcommand{\transform}[1][]{%
	\ifthenelse{\equal{#1}{}}{\mathbf{T}}{\mathbf{T}_{#1}}%
}
\newcommand{\pearl}{PE{\footnotesize A}RL}
\newcommand{\mylabel}{\ell}
\usepackage[scr=boondox]{mathalfa}
\newcommand{\outlier}{\mathscr{O}}
\newcommand{\se}{{$SE\left(3\right)$}}

\title{\LARGE \bf
Multimotion Visual Odometry (MVO):\\
Simultaneous Estimation of Camera and Third-Party Motions
}

\author{Kevin M. Judd$^{1}$ and Jonathan D. Gammell$^{1}$ and Paul Newman$^{1}$%
\thanks{$^{1}$Oxford Robotics Institute, University of Oxford, United Kingdom
        {\tt \{kjudd, gammell, pnewman\}@robots.ox.ac.uk}.}%
}

\begin{document}

\newcommand{\ORIfootertext}{This updated manuscript corrects the experimental results published in the proceedings\\of the 2018 IEEE/RSJ International Conference on Intelligent Robots and Systems (IROS).}

\maketitle
\fancyhf{}
\pagestyle{fancy}
\fancyfoot[C]{\bf \ORIfootertext}
\renewcommand\headrulewidth{0pt}
\fancypagestyle{titlepagestyle}{ %
	\fancyfoot[C]{\bf \ORIfootertext}%
	\renewcommand{\headrulewidth}{0pt} 
}

\begin{abstract}
Estimating motion from images is a well-studied problem in computer vision and robotics. Previous work has developed techniques to estimate the motion of a moving camera in a largely static environment (e.g., visual odometry) and to segment or track motions in a dynamic scene using known camera motions (e.g., multiple object tracking).

It is more challenging to estimate the unknown motion of the camera and the dynamic scene simultaneously. Most previous work requires \textit{a priori} object models (e.g., tracking-by-detection), motion constraints (e.g., planar motion), or fails to estimate the full \se{} motions of the scene (e.g., scene flow). While these approaches work well in specific application domains, they are not generalizable to unconstrained motions.

This paper extends the traditional visual odometry (VO) pipeline to estimate the full \se{} motion of both a stereo/RGB-D camera and the dynamic scene. This multimotion visual odometry (MVO) pipeline requires no \textit{a priori} knowledge of the environment or the dynamic objects. Its performance is evaluated on a real-world dynamic dataset with ground truth for all motions from a motion capture system.
\end{abstract}

\section{Introduction}

Visual navigation is an important area of research in robotics. Visual odometry (VO) estimates the motion of a camera (i.e., its egomotion) relative to observed static objects within a scene \cite{moravec1980}. These static objects must be accurately segmented from any dynamic noise, and this segmentation itself is an area of research focus \cite{matthies1987}. Less research in visual navigation has focused on also analyzing the dynamic regions of the scene that these approaches reject.

This motion estimation problem requires both \emph{estimation}, i.e., calculating the motion of a set of points, and \emph{segmentation}, i.e., clustering points according to their movement between observations. 
The interdependence of these tasks creates a \emph{chicken-and-egg} problem that is addressed in VO systems by using heuristics (e.g., number of features) to select the egomotion and ignore the other motions in the scene.
These heuristics are not readily extensible to \emph{multimotion} estimation problems
and analyzing multiple independently moving bodies remains a challenging problem for state-of-the-art vision systems.
This paper extends traditional VO to multimotion visual odometry (MVO) and applies state-of-the-art techniques to estimate trajectories for \emph{every} motion in a scene.\looseness=-1 

\begin{figure}[t]
	\centering
	\includegraphics[clip,width=0.8\columnwidth,page=1]{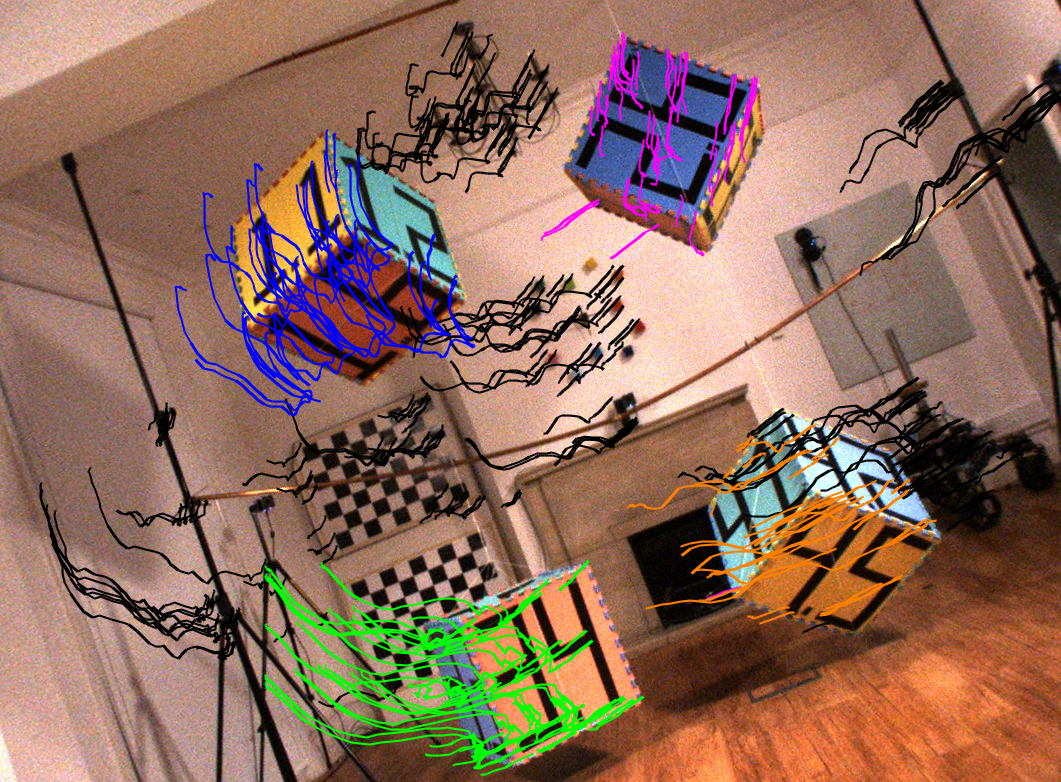}%
	\caption{Motion segmentation produced by our multimotion visual odometry (MVO) system. The egomotion of the camera is estimated from the static points in the scene shown in black. The other colors represent the segmentation of the other motions in the scene.\looseness=-1}
	\vspace{-2mm}
	\label{fig:marquee}
\end{figure}

\begin{figure*}[t]
	\centering%
	\hfill\subfloat[\label{fig:problem_full}]{%
		\includegraphics[clip,width=.95\columnwidth]{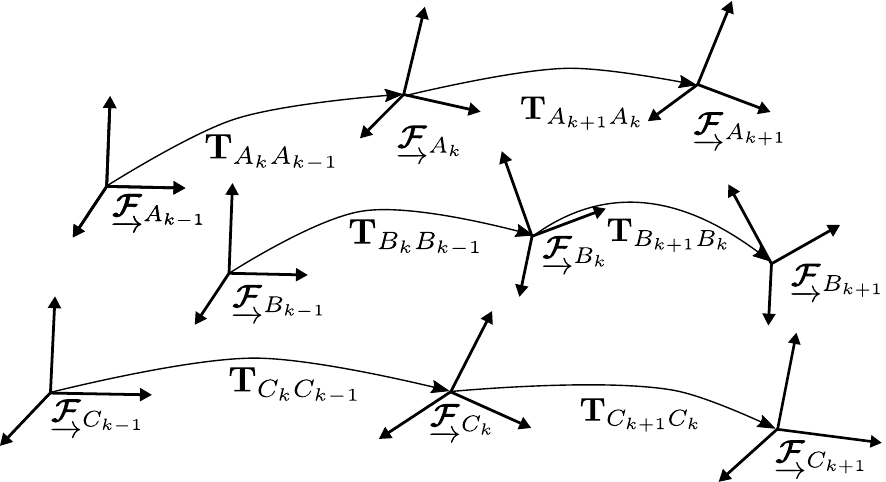}%
	}\hfill
	\subfloat[\label{fig:problem_focus}]{%
		\includegraphics[clip,width=0.73\columnwidth]{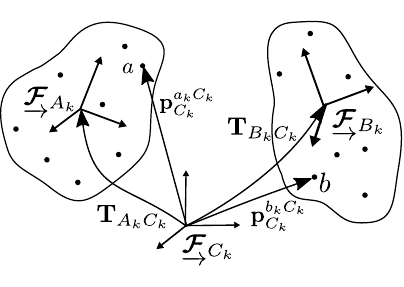}%
	}\hfill%
	\caption{Illustrations of the multimotion estimation problem showing the motion of frames through time, (a), and the relative point observations, (b). A series of two independent third-party motions, $\protect\underrightarrow{\boldsymbol{\mathcal{F}}}_{A}$ and $\protect\underrightarrow{\boldsymbol{\mathcal{F}}}_{B}$, are observed by a moving camera, $\protect\underrightarrow{\boldsymbol{\mathcal{F}}}_{C}$, through feature measurements on the objects, $\mathbf{p}^{{a_k}{C_k}}_{C_k}$ and $\mathbf{p}^{{b_k}{C_k}}_{C_k}$. Solving the problem requires simultaneously segmenting and estimating the set of measurements.}
	\vspace{-4mm}
	\label{fig:problem}
\end{figure*}

MVO applies multimodel fitting techniques (e.g., CORAL \cite{amayo2018}) to the traditional VO pipeline to simultaneously estimate the trajectories of all motions within a scene. Sparse, 3D visual features are decomposed into independent rigid motions and the trajectories of all of these motions, including the egomotion of the camera, are estimated simultaneously (\cref{fig:marquee}). This paper demonstrates MVO on a stereo camera, but the technique is applicable to a variety of other 3D sensors, including RGB-D cameras and lidar. To the best of our knowledge, this is the first approach capable of estimating the full \se{} trajectory of every rigid motion in a complex, dynamic scene from a stereo/RGB-D camera without relying on simplifying constraints or fragile initialization.

\section{Background} \label{sec:review}
The constituent aspects of the multimotion estimation problem are often referred to as multiple object tracking (MOT) \cite{milan2016} and multibody structure from motion (MBSfM) \cite{costeira1998}. As such, the majority of approaches focus primarily on one part of the problem and either do not fully estimate \se{} motions (\cref{sec:flow,sec:tracking}), depend on simplifying constraints and assumptions (\cref{sec:subspace,sec:statistical}), or require fragile initialization steps (\cref{sec:energy}).

\subsection{Problem Definition} \label{sec:problem}
Discrete multimotion estimation is the problem of estimating all the motions, including the camera, in a scene from a set of point observations at each time step. It both estimates the motions as a series of discrete \se{} transforms and associates the observed tracklets with the estimated motions.

Dynamic environments consist of the static background, a moving observer, i.e., the camera, and one or more independent, third-party motions. The pose of a motion, $\mylabel{}$, at each discrete time, $k$, is represented as a coordinate frame, $\protect\underrightarrow{\boldsymbol{\mathcal{F}}}_{\mylabel{}_k}$, and related to a privileged initial pose through an \se{} transform, $\transform[{\mylabel{}_{k}\mylabel{}_{1}}]$ (\cref{fig:problem_full}). A sequence of these transforms over a set of $K$ frames constitutes the trajectory of the motion, $T_\mylabel{} \coloneqq \left(\transform[{\mylabel{}_k,\mylabel{}_1}]\right)_{k=1\dots K}$. Likewise, a sequence of observations of a point, $j$, by a moving camera, $C$, over multiple frames forms a tracklet, $p^j \coloneqq \left(\mathbf{p}^{{j_k}{C_k}}_{C_k}\right)_{k=1\dots K}$, where $C_k$ refers to the observing camera frame at time $k$ (\cref{fig:problem_focus}). Tracklets moving with a common trajectory can be grouped into bulk motions as $\mathcal{P}_\mylabel{} \coloneqq \left\{p^j\right\}_{1\dots N}$.

Motions estimated from these measurements are \emph{egocentric}. They can be represented as \emph{geocentric} motions after identifying one motion as the camera, $T_C$.

\subsection{Flow Techniques} \label{sec:flow}
Optical flow \cite{horn1981}, scene flow \cite{vedula1999}, and sparse scene flow \cite{lenz2011} are approaches for finding the 2D or 3D velocity vector of pixels or feature points in a scene. 
These individual velocities are inherently translational and motions involving rotations (i.e., \se{} transforms) can only be estimated from segmentations of three or more velocities. 
In the presence of small rotations, these segmentations can be achieved using flow discontinuities \cite{menze2015} or the vector distance between velocities \cite{lenz2011}. \looseness=-1

\begin{figure}[t]
	\vspace{-1mm}
	\centering
	\hfill\subfloat[\label{fig:flow}]{%
		\includegraphics[clip,width=0.45\columnwidth]{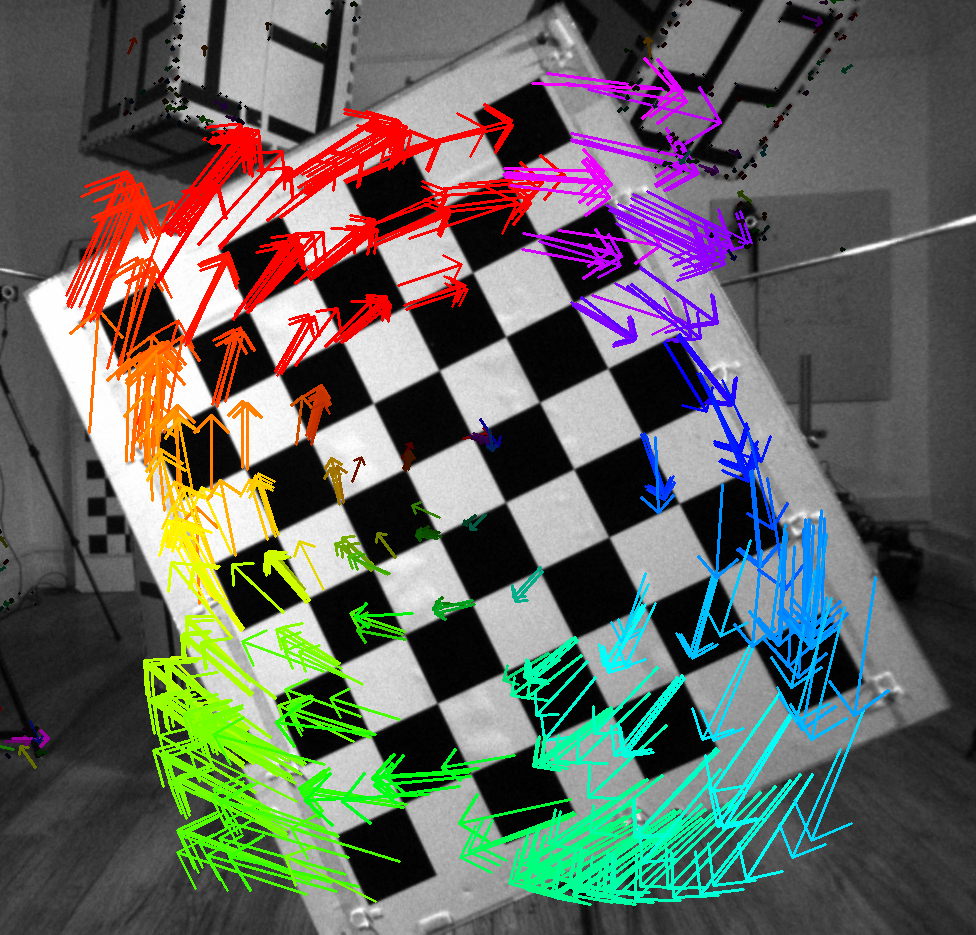}
	}\hfill
	\hfill\subfloat[\label{fig:mvo}]{%
		\includegraphics[clip,width=0.45\columnwidth]{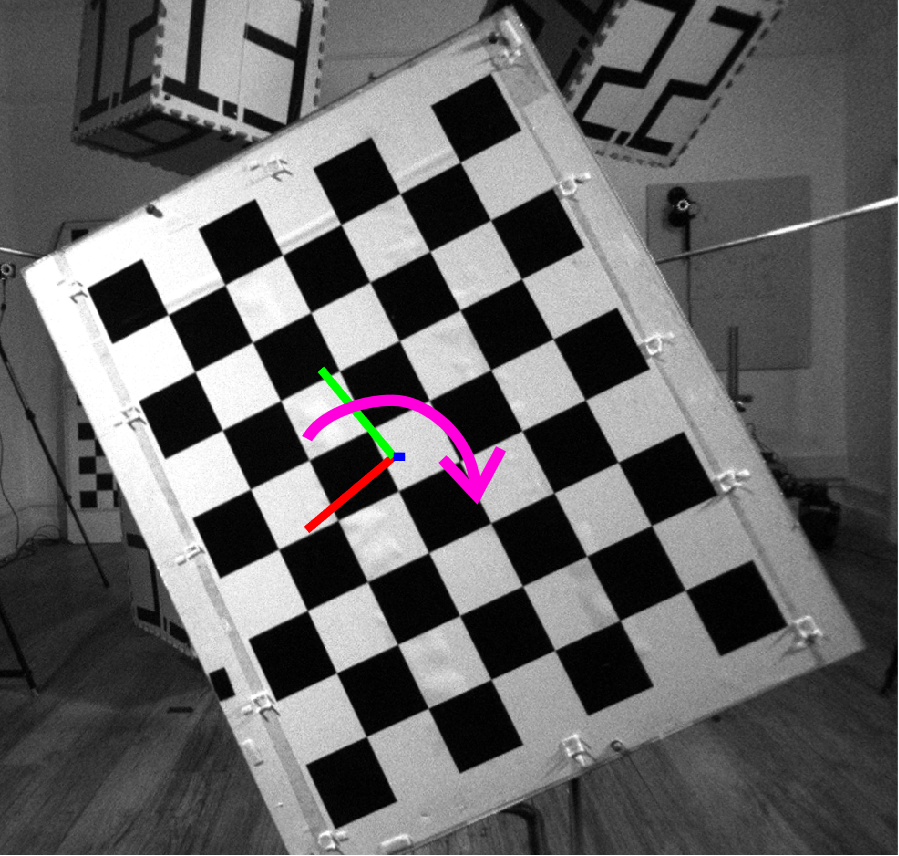}
	}\hfill
	
	\caption{Scene flow \cite{lenz2011}, (a),  and MVO, (b), on a checkerboard rotating about the optical axis. Flow techniques inherently calculate individual translational velocities. Since each point on a rotating body has a different tangential velocity, flow techniques fail to segment the velocities into bulk motions. MVO estimates motions as SE(3) transforms even in the presence of rotation. To estimate these motions, flow techniques must solve an equivalent segmentation and estimation problem in the space of velocities.}
	\vspace{-3mm}
	\label{fig:rotation_comparison}
\end{figure}

Larger rotations result in smoothly varying tangential velocities that provide no clear segmentations  (\cref{fig:rotation_comparison}a). 
To correctly estimate these motions, flow techniques must solve an equivalent segmentation and estimation problem posed in the space of velocities. 
In contrast to these flow techniques, MVO simultaneously segments and estimates full \se{} transforms of motions in the scene (\cref{fig:rotation_comparison}b).

\subsection{Tracking-by-Detection Methods} \label{sec:tracking}
Appearance-based tracking techniques detect objects in images and then solve the association and motion estimation problems \cite{milan2016}. Kalman or particle filters are widely used to estimate the motion of the detected objects given a motion model but struggle to handle detection errors or occlusions \cite{bar2004}. \looseness=-1 

Byeon et al. \cite{byeon2018} propose an optimization framework for tracking multiple objects and estimating their trajectories from multiple static cameras by incorporating reconstruction and motion dynamics in their cost function. Zhang et al. \cite{zhang2008} model tracking as a mininimum-flow problem on a graph where nodes represent detections and edges represent transitions between frames. 

Object detectors are designed either for some specified class of objects or dynamically for some object of interest. These detectors are therefore specialized for a specific set of applications or are fragile to appearance changes and need to be refined over time \cite{kalal2012}. 

Many of these approaches require either static or known camera motion and therefore need to incorporate separate egomotion estimators \cite{wang2007}.
The object positions only exist in $\mathbb{R}^3$ and do not fully encapsulate the \se{} motions of the objects. Kundu et al. \cite{kundu2011} extend egomotion estimation with MBSfM techniques similar to \cite{wang2007} to estimate the \se{} trajectories of the third-party motions in a scene, but they constrain all the motions to the horizontal plane.

Unlike these appearance-based techniques, MVO relies on low-level feature tracking. This means it can handle large changes in object appearance over time so long as a suitable number of features remain stable between each pair of frames. MVO also estimates the full, unconstrained \se{} motion within a scene, including the egomotion of the camera.

\subsection{Subspace Methods} \label{sec:subspace}
Subspace techniques cluster sparse feature points and their \se{} motions into lower-dimensional subspaces using the affine camera model. The affine model approximates the nonlinear perspective projection with a linear parallel projection. This simplifies the camera model but introduces severe projection errors in scenes with a wide field of view or a large depth of field \cite{hartley2003}.

Tomasi and Kanade \cite{tomasi1992} use the affine model and matrix factorization to decompose tracked image points into a motion and a shape matrix. Costeira and Kanade \cite{costeira1998} extend this technique to mutiple bodies where points may belong to different objects. This approach is inherently fragile as noise propagates through the factorization in complex ways \cite{kanatani2001}. It also requires points to be tracked for the entirety of the estimation window, which is difficult in complex scenes. 

This formulation was extended by using an optimization framework to allow feature point dropouts \cite{vidal2004} and by merging motions to mitigate the effect of noise \cite{kanatani2001, wu2001}. 
While these techniques are able to estimate full \se{} motions, they still depend upon an affine camera model, meaning they fail under any significant perspective effects. 

These factorization techniques have also been extended to the perspective camera model by estimating depth in a preprocessing stage \cite{li2007} or by applying geometric constraints \cite{vidal2003} but still remain very sensitive to noise. In comparison, MVO can robustly estimate full \se{} trajectories using a perspective model while also handling the significant feature tracking failures characteristic of dynamic scenes.

\subsection{Sampling Methods} \label{sec:statistical}
Sampling methods estimate and fit models (e.g., motion trajectories) to a subset of the data before evaluating them across its entirety. RANSAC \cite{fischler1981} is a popular framework to fit a model to data in the presence of noise. Points are sampled from the data to estimate a hypothesis model, which is then used to segment the data into inliers and outliers according to their fit. Hypotheses are repeatedly generated for some number of iterations and the model with the most inliers is selected as the segmentation and estimation.

\begin{figure}
	\centering
	\hfill\subfloat[\label{fig:num_models_found_line}]{%
		\includegraphics[height=26mm]{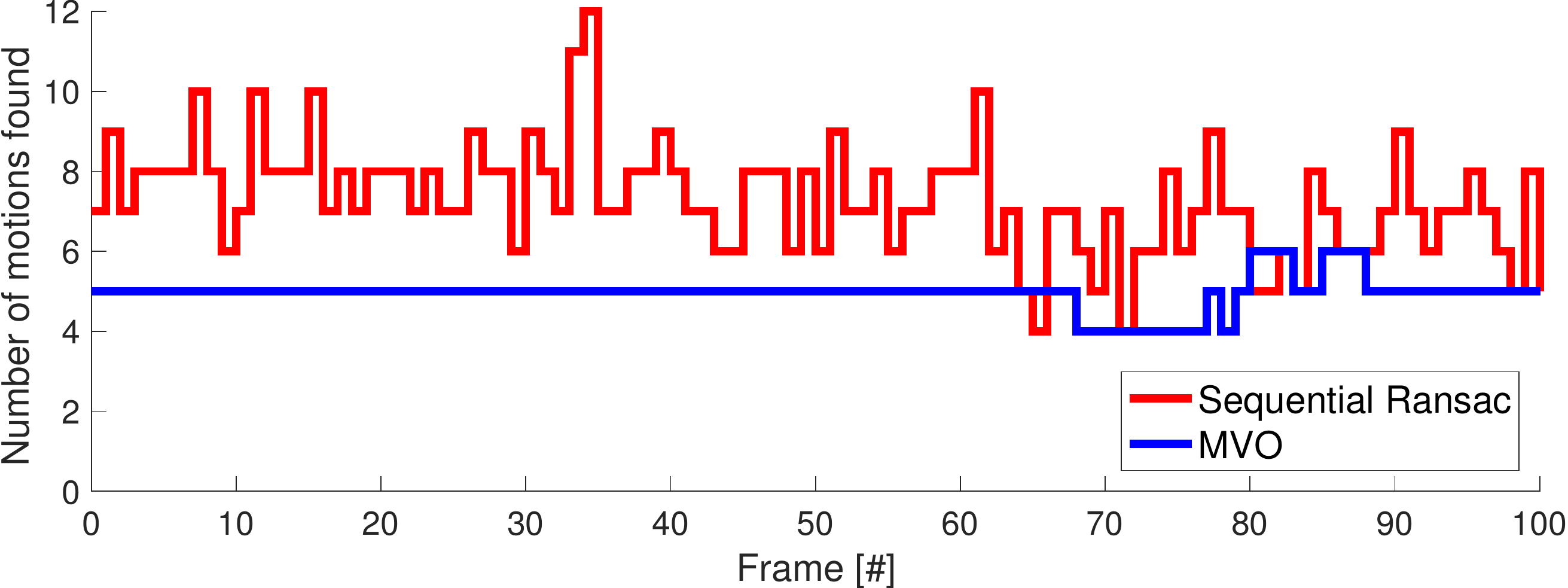}
	}
	\subfloat[\label{fig:num_models_found_hist}]{%
		\includegraphics[height=26mm]{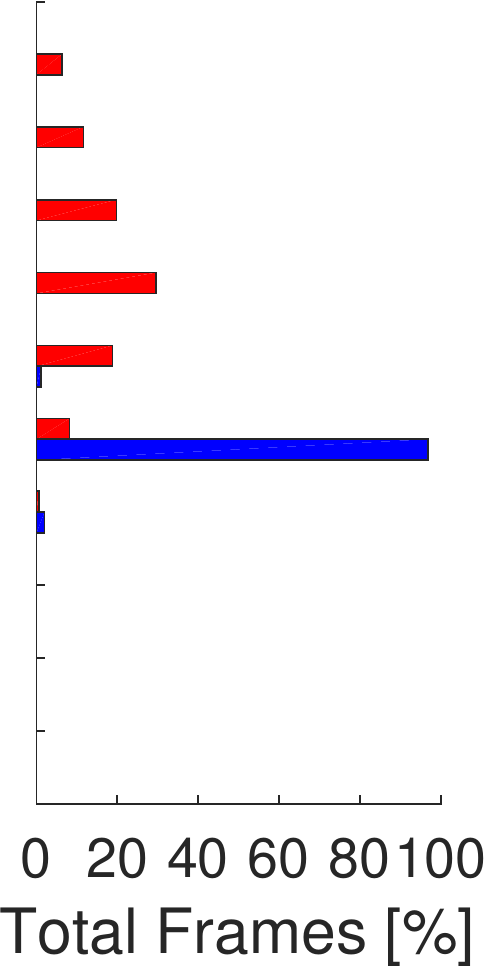}
	}\hfill
	\caption{A comparison of the number of models found by a sequential RANSAC (red) and MVO (blue) over 500 frames of a five-motion experiment (\cref{sec:experiments}). The first 100-frame section of the sequence is shown in (a) with the cumulative totals for the full sequence in (b). Without prior knowledge of the number of motions, sequential RANSAC greedily overfits to noise and finds an inconsistent number of models. MVO consistently segments the correct number of motions, except for frames 68--87 when some of the motions partially move outside the camera frustum. While RANSAC only found the correct number of models 31.2\% of the time, MVO found the correct number 96.8\% of the time. }
	\vspace{-3.5mm}
	\label{fig:num_models}
\end{figure}

Torr \cite{torr1998} extends the RANSAC framework to multiple models by finding the dominant model, removing those points that fit the model, and then recursively applying RANSAC to the remaining points.
This recursive, sequential RANSAC framework is efficient at finding the dominant models in a scene, but the ability to sample consistent models decreases as models are removed and the signal-to-noise ratio of the remaining points decreases. Sabzevari and Scaramuzza \cite{sabzevari2016} apply geometric and kinematic constraints to reduce the required number of points to estimate a motion model and then realign point assignments to the best set of motion hypotheses in a separate step. They use the same matrix formulation as \cite{li2007}, meaning points must be successfully tracked through the entirety of the window, and their applications are limited by the constraints.\looseness=-1

Other techniques \cite{schindler2006, ozden2010} use sampling methods to generate a large number of initial model hypotheses, realizing many of them would be redundant or poorly fit the data. Models are merged if their inlier sets are largely overlapping, and the models with the largest nonoverlapping inlier sets after merging are taken as the constituent scene motions.

These sampling methods are efficient but the probability of sampling inliers all from a single model decreases rapidly with the signal-to-noise ratio. 
Finding a motion in complex dynamic scenes is challenging because all other motions are outliers that decrease the signal-to-noise ratio and make it harder to find correct models.
As a result, many of these sampling-based initializations struggle to find correct models.\looseness=-1 

Without prior knowledge of the number of models in the data, RANSAC tends to greedily overfit to noise and finds erroneous or incomplete models (\cref{fig:num_models}). In contrast, MVO estimates all models simultaneously and requires no \emph{a priori} knowledge of the number of models.

\begin{figure*}[t]
	\centering
	\def\svgwidth{\textwidth}
	\includegraphics[width=\textwidth]{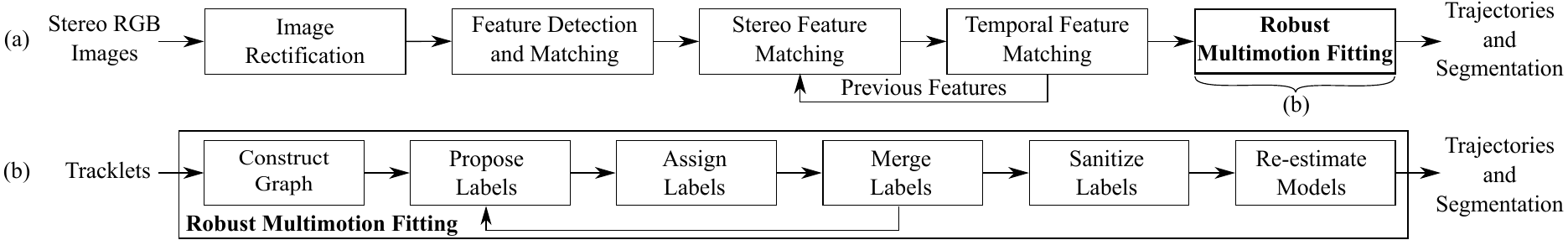}
	\caption{An illustration of the stereo multimotion visual odometry (MVO) pipeline, (a), including details of the multimotion fitting component, (b). The pipeline operates on RGB stereo image pairs and estimates the segmentation and trajectories of all motions in the scene. It extends the standard stereo visual odometry pipeline by replacing the egomotion estimator with a multimotion estimator that operates on the tracklets produced by the previous stages. The multimotion-fitting pipeline builds a neighborhood graph based on the distances between pairs of points throughout the estimation batch. The pipeline iteratively splits and estimates new labels using the neighborhood graph, performs label-assignment using CORAL \cite{amayo2018}, and merges labels that can be considered redundant until convergence. Finally, the label set is sanitized and a full-batch estimation step produces the final segmentation and trajectories.}
	\vspace{-3.5mm}
	\label{fig:pipeline}
\end{figure*}

\subsection{Energy Minimization Methods} \label{sec:energy}
Energy minimization approaches segment data into multiple labels simultaneously by reducing a cost function. In multimotion estimation, this cost is designed to encompass how well the estimated trajectories describe the data, e.g., reprojection or photoconsistency error, as well as encourage piecewise smoothness throughout the scene. Smoothness is enforced over a graph structure, usually either a dense Markov random field \cite{nieuwenhuis2013} or a sparse, feature-based graph \cite{isack2012}. 

Rother et al. \cite{rother2004} use a minimization framework to find a binary segmentation of the static background from a manually selected dynamic foreground object, but the approach can only segment a single object within a bounding box. 
\pearl{} \cite{isack2012} uses $\alpha$-expansion and model refitting to iteratively estimate both models and point assignments in an expectation-maximization framework. The framework can be applied to motion segmentation by first sampling the data to estimate a large number of motion models (similarly to \cite{schindler2006, ozden2010}) and then refining the models and segmentation with \pearl{}. Roussos et al. \cite{roussos2012} use \pearl{} as part of a dense expectation-maximization pipeline that estimates depth, motion, and segmentation from monocular images in an offline manner. Optical flow is used to initialize the depth maps and the approach is crucially dependent on this initialization.

R\"unz and Agapito \cite{runz2017} use a similar optimization framework to segment dense RGB-D camera observations. Using this segmentation they create multiple 3D object models whose \se{} motions are then tracked, establishing new motions online. This approach requires an initialization phase that seeds the structure and segmentation of the background of the scene.\looseness=-1
	
These approaches are capable of estimating full \se{} motions but are dependent upon comprehensive initialization, often using RANSAC, as they are designed around refining existing labels. In comparison, MVO iteratively proposes and refines labels, allowing it to find motions that may be difficult to initially segment with sampling methods alone.

\subsection{Multimotion Visual Odometry (MVO)}
To the best of our knowledge, this paper introduces the first online approach capable of directly estimating full \se{} trajectories for every motion in a complex, dynamic scene from a stereo/RGB-D camera using only a rigid-body assumption. The approach uses multilabeling and estimation techniques to model the motions of tracklet features over multiple frames. The hypothesis trajectories are then applied to individual tracklets by CORAL \cite{amayo2018}, a convex optimization approach to the multilabeling problem. These hypotheses are iteratively improved through splitting and merging of the models, unlike other labeling approaches that initially sample them from the scene. The full \se{} trajectory of \emph{each} motion is finally estimated using traditional VO batch estimation techniques. This approach is evaluated on a dataset containing ground-truth trajectories for all motions in the scene. \looseness=-1

\section{Methodology} \label{sec:methodology}
The stereo MVO pipeline (\cref{fig:pipeline}a) extends the traditional stereo VO pipeline to \emph{multimodel} segmentation and estimation. The incoming RGB stereo images are first rectified and undistorted according to known camera extrinsics and intrinsics. Salient image points are detected and matched across left and right images in each stereo pair and across temporally consecutive pairs of stereo frames. These stereo- and temporally-matched feature points are back-projected into the 3D space using the camera intrinsics, forming world- and image-space tracklet histories for each feature point. This set of tracklets, $\mathcal{P} \coloneqq \left\{p\right\}$, becomes the input to the multimotion segmentation and estimation engine (\cref{fig:pipeline}b). These tracklets could alternatively be found by associating observations from other 3D sensors (e.g., RGB-D cameras) over time.

The multimotion engine segments tracklets by their observed motion, which is a combination of camera and object motions. In the absence of \emph{a priori} information about the scene, each group of tracklets is used to estimate a camera egomotion by assuming those tracklets belong to a static object. These camera egomotion \emph{hypotheses} can later be converted into estimates of the camera and object motions by identifying the static part of the scene (e.g., as in VO).

The segmentation and estimation are posed as a multilabeling problem where a a label, $\mylabel{}$, represents the egomotion hypothesis, $^\mylabel{} T_C$, calculated from a group of tracklets, $\mathcal{P}_\mylabel{} \subseteq \mathcal{P}$.
These labels are assigned by minimizing a cost function over a graph of all observed tracklets (\cref{sec:neighborhood}).  
New labels are proposed for each disconnected component of a label's subgraph through a multiframe RANSAC procedure (\cref{sec:splitting}). 
Motion labels are assigned to minimize the reprojection residual of the associated trajectory and maximize the label smoothness in the graph (\cref{sec:assignment}). 
An outlier label, $\outlier{}$, is assigned to points whose motions are not well explained by any other label.
Redundant and oversegmented labels are then merged (\cref{sec:merging}). \looseness=-1 

The algorithm iterates this process until label convergence. The final labels are then sanitized and any remaining outliers are rejected (\cref{sec:sanitization}) before a final, full-batch estimation of each label (\cref{sec:batch}). Egocentric or geocentric trajectories are found by selecting a label to represent the motion of the camera (\cref{sec:geocentric}).\looseness=-1

\subsection{Graph Construction} \label{sec:neighborhood}
The rigid-body assumption is approximated through a geometric neighborhood graph, $\graph$. Each vertex of the graph represents an observed tracklet and is connected to its $k$-nearest-neighbors. The distance between two vertices is defined as the maximum distance in image space between those image tracklets over the entire batch,
\begin{equation*}
d\left(p^i,p^j\right) \coloneqq \max\limits_{k=1 \dots K} \norm{\mathbf{s}\left(\mathbf{p}_{C_{k}}^{{i_{k}}{C_{k}}}\right) - \mathbf{s}\left(\mathbf{p}_{C_{k}}^{{j_{k}}{C_{k}}}\right)}_2,
\end{equation*}
where $\mathbf{s}\left(\cdot\right)$ applies the nonlinear perspective camera projection. This allows for edges between features that are consistently close while not connecting features that are ever far apart or that never coexist in a frame. This connectivity forms the basis for label generation and assignment.\looseness=-1 

\subsection{Label Proposal} \label{sec:splitting}
The label set, $\mathcal{L}$, must dynamically grow and adapt to correctly converge in a given scene. To accomplish this, new labels are generated by splitting label support groups whenever their tracklets' motions could more accurately be explained by multiple trajectories. A potential new label, $\mylabel{}'$, is generated for each fully-disjoint component of the subgraph defined by the label's support, $\graph_{\mylabel{}'}\subseteq\graph_\mylabel{}\subseteq\graph$. This ensures a level of spatial smoothness while allowing new labels to be proposed from large label supports comprised of tracklets from spatially or temporally distinct motions in the scene. 

The new label proposals are generated by computing both the dominant motion of the given points and the segmentation between inliers and outliers of that motion. 
This single-motion segmentation and estimation problem is solved by applying RANSAC in a frame-to-frame fashion, similar to standard VO systems. \looseness=-1

Three tracklets are sampled from those visible in the current, $k$, and previous, $k-1$, frames to estimate the $SE(3)$ transform between the two frames ${^{\mylabel{}'} \transform[{{C_{k}},{C_{k-1}}}]}$. The proposed transform is evaluated according to how many tracklet reprojection residuals, 
\begin{equation} \label{eq:datacost}
e_k\left(p^{j}, {^{\mylabel{}'} T_C}\right) \coloneqq \norm{\mathbf{s}\left(\mathbf{p}_{C_{k}}^{{j_{k}}{C_{k}}}\right) - \mathbf{s}\left({^{\mylabel{}'}\transform[{{C_{k}}{C_{k-1}}}]}\mathbf{p}_{C_{k-1}}^{{j_{k-1}}{C_{k-1}}}\right)}_2,
\end{equation}
are within a given threshold error, $e_{\mathrm{th}}$. This process is repeated many times and the transform with the largest inlier set is appended to the proposed trajectory hypothesis, ${^{\mylabel{}'}T_C}$.\looseness=-1 

Any tracklets found to be outliers of the newly estimated models are appended to the outlier label, $\outlier{}$. New labels are generated from the outlier label last.

\subsection{Label Assignment} \label{sec:assignment}
Each tracklet, $p\in\mathcal{P}$, is assigned a label, $\mylabel{}\in\mathcal{L}$, to minimize the energy functional,
\begin{equation} \label{eq:coral}
\begin{aligned}
E\left(\mathcal{L}\right) \coloneqq& \underbrace{\sum_{p\in\mathcal{P}} \rho\left(p, \mylabel{}\left(p\right)\right)}_\text{Residual} + \underbrace{\lambda\sum_{\left(p,q\right)\in\graph} \omega_{pq}V\left(p,q\right)}_\text{Smoothness} \\ 
&+ \underbrace{\sum_{\mylabel{}\in\mathcal{L}} \gamma_{\mylabel{}}\psi_{\mylabel{}}}_\text{Complexity}
\end{aligned},
\end{equation}
where $\mylabel{}\left(p\right)$ gives the label currently assigned to $p$. The energy functional combines the residual error, the label smoothness, and the label complexity term, using a user-selected proportionality parameter, $\lambda$. 

\paragraph{Residual}
The residual term penalizes labels that poorly describe the observed data. It is defined as the sum of the residual errors of applying the label trajectories to tracklets. The residual for each point-label pair is defined as
\begin{equation*}
\rho\left(p^j, \mylabel{}\right) \coloneqq \max\limits_{k \in 1\dots K} e_k\left(p^j,{^{\mylabel{}}T_C}\right),
\end{equation*}
where $e_k$ as defined in \cref{eq:datacost}. 

\paragraph{Smoothness}
The smoothness term penalizes neighboring tracklets that do not share the same label by an edge cost, $\omega_{pq}$. This encourages a piecewise-smooth solution. It is a weighted sum of all edges penalized according to\looseness=-1 
\begin{equation*}
V\left(p,q\right) \coloneqq 
\begin{cases}
1 & \mylabel{}\left(p\right) \neq \mylabel{}\left(q\right) \\
0 & \mathrm{otherwise}
\end{cases}.
\end{equation*}

\paragraph{Complexity}
The complexity term encourages a compact solution by penalizing the use of many labels. It is the sum of the per-label cost, $\gamma_{\mylabel{}}$, of each label with non-empty support set according to the function,
\begin{equation*}
\psi_{l} \coloneqq
\begin{cases}
1 & \vert\mathcal{P}_\mylabel{}\vert > 0 \\
0 & \mathrm{otherwise}
\end{cases}.
\end{equation*}

\paragraph{Outliers}
The outlier label, $\outlier{}$, is designed to be attractive to all points whose motions are not well explained by  existing labels. The residual energy of the outlier label decays exponentially with that of the best-fitting label, 
\begin{equation*}
\rho\left(p^j, \outlier{}\right) \coloneqq \alpha\exp\left(-\frac{\min\limits_{\mylabel{} \in \mathcal{L}}\rho\left(p^j, \mylabel\right)}{\beta}\right),
\end{equation*}
where $\alpha$ and $\beta$ are tuning parameters. Points that are well-explained by an extant label will have high outlier data cost. The label cost, $\gamma_{\outlier{}}$, for the outlier label is zero as outliers are assumed to always exist.

Given the current label set, $\alpha$-expansion (e.g., \pearl{} \cite{isack2012}) or convex optimization (e.g., CORAL \cite{amayo2018}) assigns a label to each tracklet to minimize the residual and smoothness energies of \cref{eq:coral}. The minimization can result in an oversegmentation due to outliers and poorly estimated intermediate trajectories. Model merging is therefore used to improve the motion estimation. \looseness=-1

\subsection{Label Merging} \label{sec:merging}
Two labels, $\mylabel{}$ and $\mylabel{}'$, may be merged if relabeling all~$\mathcal{P}_{\mylabel'}$~as $\mylabel{}$ would decrease the total energy of \cref{eq:coral}. This occurs when the increase in residual error due to reduced overfitting is less than the cost of using the label, $\gamma{}_{\mylabel{}'}$, and any change in smoothness.\looseness=-1

Only the periods during which the two labels' supports overlap are considered because there is no cost for applying a new label to portions of the batch in which the tracklets do not exist. When more than one merge would reduce the total energy, the one that results in the greatest decrease in cost is chosen. Merging continues until no more merges would reduce \eqref{eq:coral}. The outlier label, $\outlier$, is excluded from merging.

The merging stage only considers label pairs with tracklets adjacent in $\graph$, i.e., those where $\graph_\mylabel{}$ is connected to $\graph_{\mylabel{}'}$. If they are disconnected, merging the two supports would be undone by the splitting routine (\cref{sec:splitting}). If the two support sets are connected then the new label will persist until the next labeling stage.

The algorithm iterates the label splitting, assignment, and merging (\crefrange{sec:splitting}{sec:merging}) until the labels converge or a maximum number of iterations have been reached. The final label set is then sanitized (\cref{sec:sanitization}) before being used to estimate the final trajectory hypotheses (\cref{sec:batch}).

\subsection{Label Sanitization} \label{sec:sanitization}
The final labels are sanitized to refine the segmentation output and remove noisy tracklets before the final model estimation. A merging step first combines any redundant labels regardless of graph connectivity as there is no subsequent splitting stage. After merging, any label with fewer than a minimum number of support tracklets or that exists for fewer than a minimum number of frames is merged with $\outlier{}$. Likewise, tracklets whose residual error is greater than a threshold, $e_{\mathrm{th}}$, are relabeled as outliers. This provides a consistent set of tracklets for the batch estimation of each motion.\looseness=-1 

\subsection{Final Model Estimation} \label{sec:batch}
For each label, an egomotion hypothesis, ${^\mylabel{}}T_C$, is estimated to explain the motion of the tracklets, $\mathcal{P}_\mylabel$, using bundle adjustment. This paper follows the single-motion approach described by Barfoot \cite{barfoot2017} to estimate the trajectory of each label in an egocentric frame. 

The system state, $\mathbf{x}$, of each label is defined to include both the estimated pose transforms, ${^\mylabel{}T} \coloneqq \left({^\mylabel{}\transform[{C_{k}C_{1}}]}\right)_{k=2\dots K}$, and the landmark points, $\left\{\mathbf{p}_{C_1}^{{j_1}{C_1}}\right\}_{j=1\dots\vert\mathcal{P}_\mylabel{}\vert}$. The state $\mathbf{x}_{jk} \coloneqq \left\{{^\mylabel{}\transform[{C_{k}C_{1}}]}, \mathbf{p}_{C_1}^{{j_1}{C_1}}\right\}$ is defined for each pair of transforms and points belonging to label $\mylabel{}$.

Each observation, $\mathbf{y}_{jk}$, of point $\mathbf{p}^{j}$ at pose ${^\mylabel{}\transform[{C_{k}C_{1}}]}$ is modeled as
\begin{equation*}
\begin{aligned}
\mathbf{y}_{jk} &\coloneqq \mathbf{g}\left(\mathbf{x}_{jk}\right) + \mathbf{n}_{jk} = \mathbf{s}\left(\mathbf{z}\left(\mathbf{x}_{jk}\right)\right) + \mathbf{n}_{jk} \\
&= \mathbf{s}\left({^\mylabel{}\transform[{C_{k}C_{1}}]}\mathbf{p}_{C_1}^{{j_1}{C_1}}\right) + \mathbf{n}_{jk}.
\end{aligned}
\end{equation*}
The measurement model, $\mathbf{g}\left(\cdot\right)$, encompasses both the motion model, $\mathbf{z}\left(\cdot\right)$, which applies $SE(3)$ transforms to observed points, and the sensor model, $\mathbf{s}\left(\cdot\right)$, derived from the perspective camera model. The model assumes additive Gaussian noise, $\mathbf{n}_{jk}$, with zero mean and covariance $\mathbf{R}_{jk}$. The least-squares cost function is defined as the difference between the measurement model and the observations,
\begin{equation*}
J \coloneqq \frac{1}{2}\sum_{jk}\mathbf{e}_{y,jk}\left(\mathbf{x}\right)^{T}\mathbf{R}_{jk}^{-1}\mathbf{e}_{y,jk}\left(\mathbf{x}\right),
\end{equation*}
where,
\begin{equation*}
\mathbf{e}_{y,jk}\left(\mathbf{x}\right) \coloneqq \mathbf{y}_{jk} - \mathbf{s}\left({^\mylabel{}\transform[{C_{k}C_{1}}]}\mathbf{p}_{C_1}^{{j_1}{C_1}}\right).
\end{equation*}

This cost is linearized about an operating point, $\mathbf{x}_{\mathrm{op}}$, and then minimized using Gauss-Newton. The operating point is perturbed according to the transform perturbations, $\{\boldsymbol{\epsilon}_k \in \mathbb{R}^{6}\}$, and landmark perturbations, $\{\boldsymbol{\zeta}_j \in \mathbb{R}^{3}\}$, which together form the full state perturbation, $\delta\mathbf{x}$. An indicator matrix $\mathbf{P}_{jk}$ is defined such that $\delta\mathbf{x}_{jk} = \mathbf{P}_{jk}\delta\mathbf{x}$. See \cite{barfoot2017} for more detail.

\begin{figure*}[tp]
	\centering
	\subfloat[Top-left swinging box]{
		\centering
		\includegraphics[clip,width=.98\columnwidth,page=1]{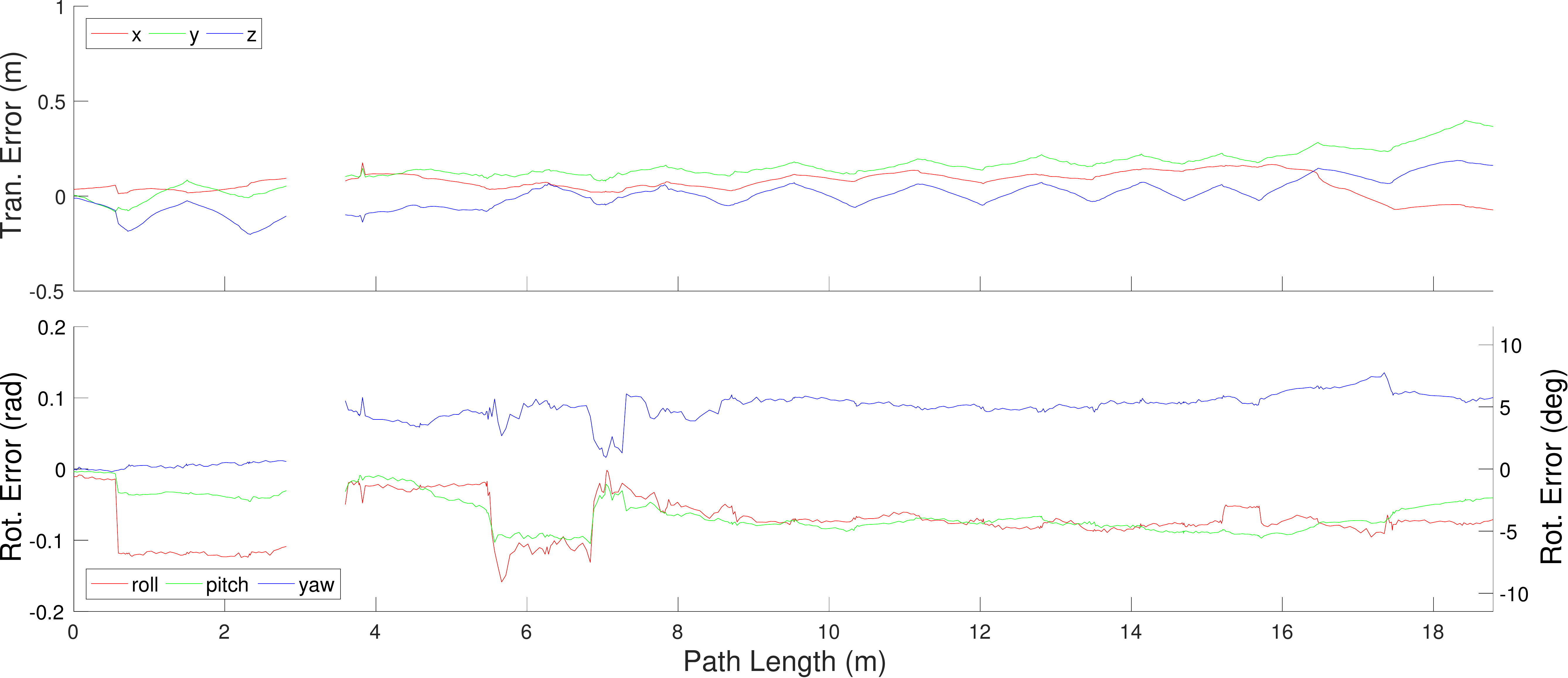}%
		\label{fig:results:topleft}
	}\hfill
	\subfloat[Top-right swinging and rotating box]{
		\centering
		\includegraphics[clip,width=.98\columnwidth,page=1]{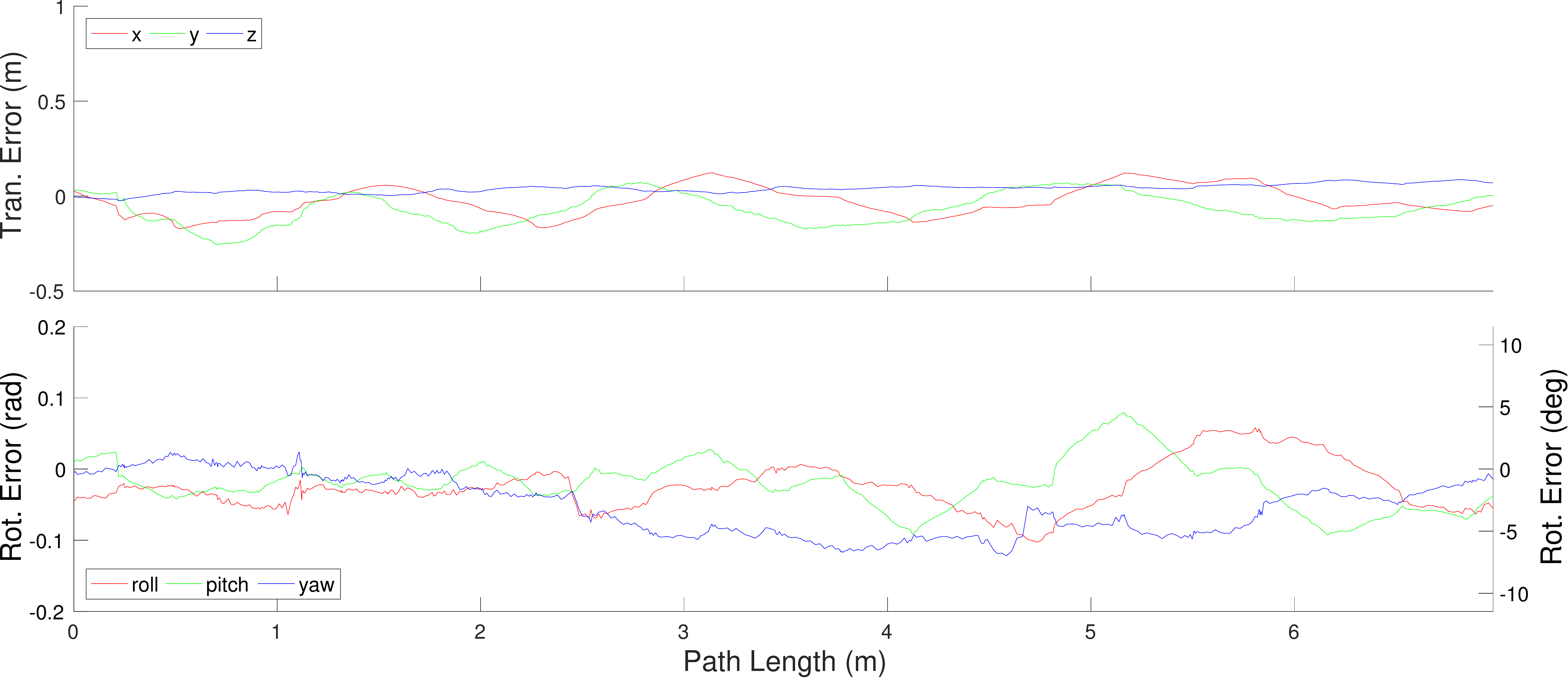}%
	}\vspace{-3mm}
	\subfloat[Bottom-left swinging box]{
		\centering
		\includegraphics[clip,width=.98\columnwidth,page=1]{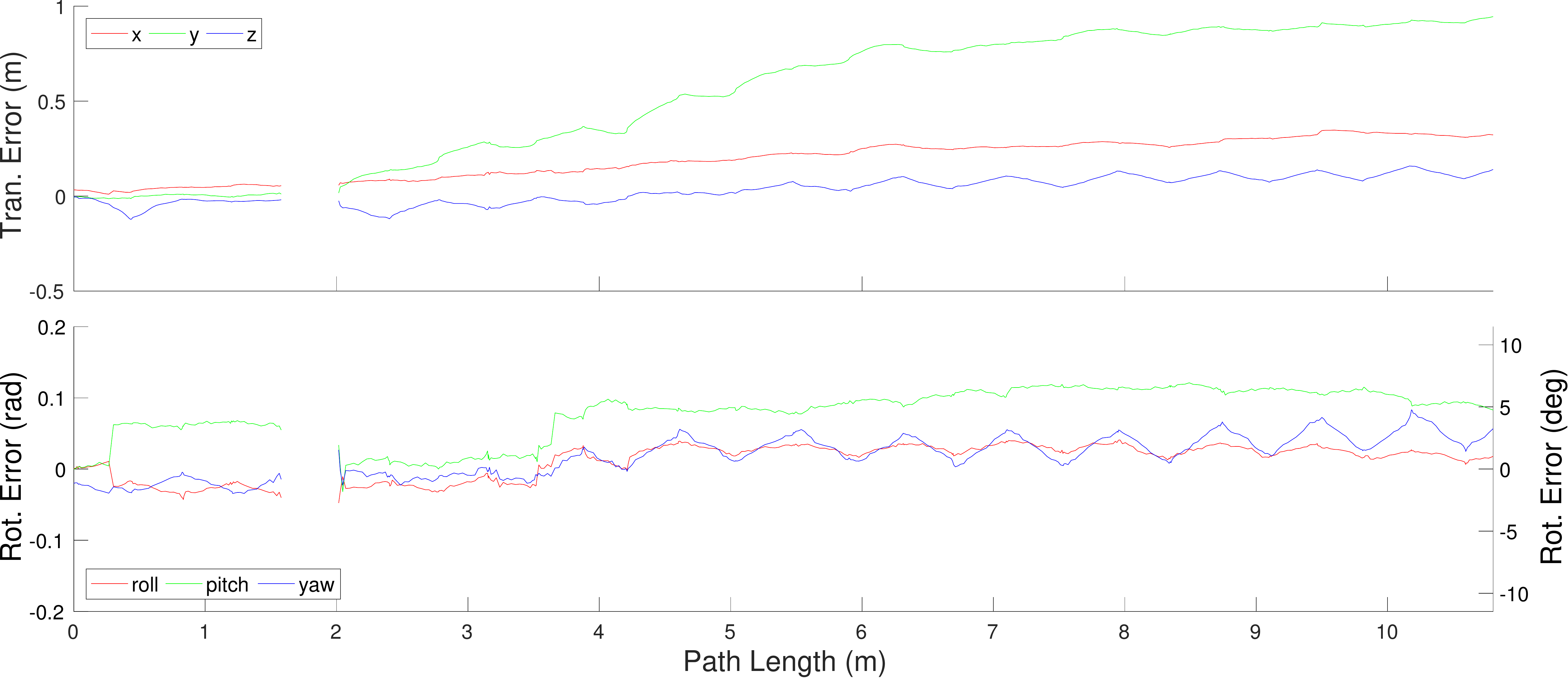}%
	}\hfill
	\subfloat[Bottom-right rotating box]{
		\centering
		\includegraphics[clip,width=.98\columnwidth,page=1]{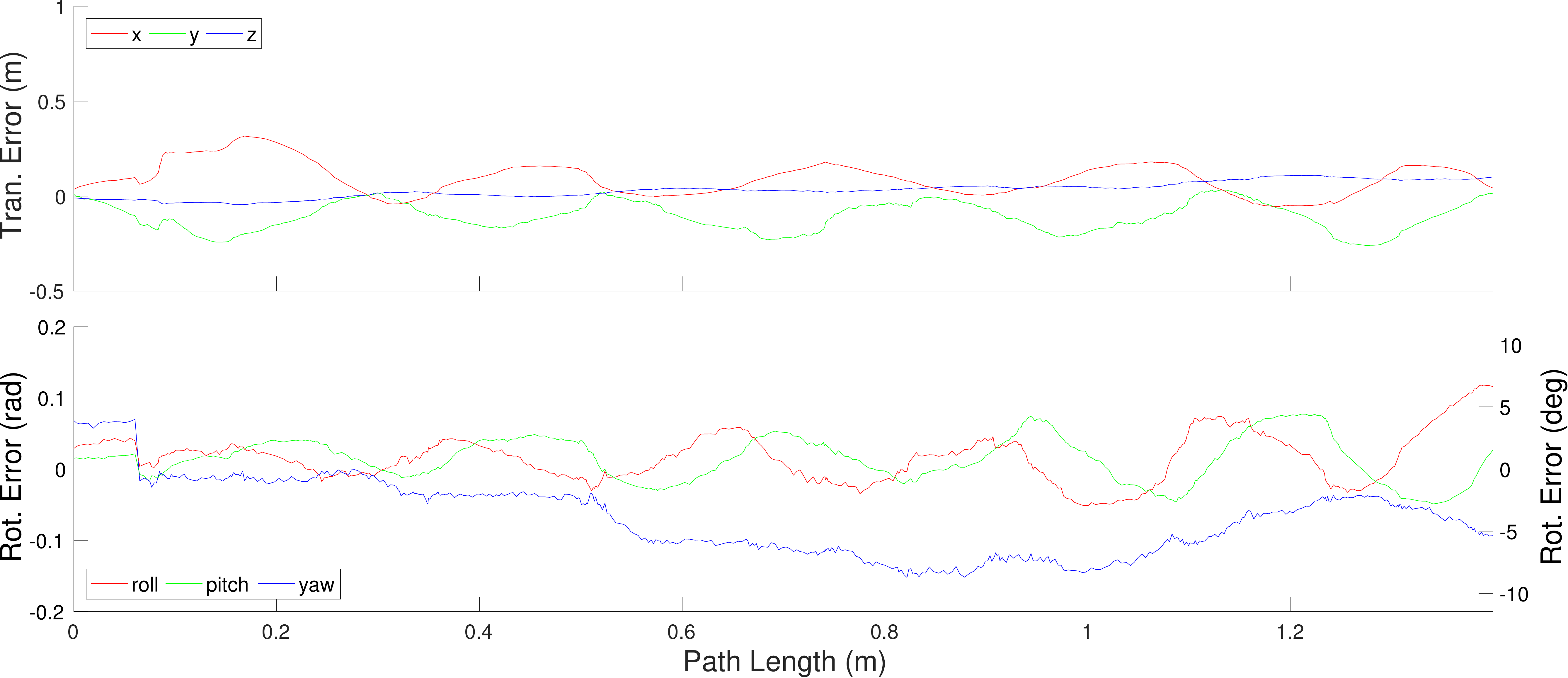}%
		\label{fig:results:rotating}
	}
	\caption{The translational and rotational errors for the estimated motion of each object over the length of the estimation window as compared to ground-truth trajectory data. Errors are reported in an arbitrary geocentric frame with the z-axis up, and the arbitrary x- and y-axes. Both the top-left block (a) and the bottom-left block (c) partially left the camera frustum near the beginning of the segment which resulted in gaps in the trajectory estimates for those blocks.}
	\label{fig:results}
\end{figure*}

The error function is linearized using $\mathbf{G}_{jk}$, the Jacobian of the measurement function, $\mathbf{g}\left(\cdot\right)$, 
\begin{equation*}
\begin{aligned}
\mathbf{G}_{jk} &\coloneqq \mathbf{S}_{jk}\mathbf{Z}_{jk}, \\
\mathbf{S}_{jk} &= \frac{\partial\mathbf{s}}{\partial\mathbf{z}}\bigg\rvert_{\mathbf{z}(\mathbf{x}_{\mathrm{op},jk})},\\
\mathbf{Z}_{jk} &= \begin{bmatrix}
\left({^\mylabel{}\transform[{\mathrm{op},{C_k}{C_1}}]}\mathbf{p}_{\mathrm{op},C_1}^{{j_1}{C_1}}\right)^{\odot} & {^\mylabel{}\transform[{\mathrm{op},{C_k}{C_1}}]}\mathbf{D}
\end{bmatrix},\\
\mathbf{D} &= \begin{bmatrix}
\mathbf{1}\\
\mathbf{0}^T
\end{bmatrix},
\end{aligned}
\end{equation*}
where the matrix operator $\left(\cdot\right)^{\odot}$ is defined in \cite{barfoot2017}. The cost function can then be linearized using 
\begin{equation*}
\begin{aligned}
J &\approx J(\mathbf{x}_{\mathrm{op}}) - \mathbf{b}^T\delta\mathbf{x}_{jk} + \frac{1}{2}\delta\mathbf{x}_{jk}^T\mathbf{A}\delta\mathbf{x}_{jk},\\
\mathbf{b} &= \sum_{jk}\mathbf{P}_{jk}^{T}\mathbf{G}_{jk}^{T}\mathbf{R}_{jk}^{-1}\mathbf{e}_{y,jk}\left(\mathbf{x}_\mathrm{op}\right),\\
\mathbf{A} &=  \sum_{jk}\mathbf{P}_{jk}^{T}\mathbf{G}_{jk}^{T}\mathbf{R}_{jk}^{-1}\mathbf{G}_{jk}\mathbf{P}_{jk}.
\end{aligned}
\end{equation*}
The optimal perturbation, $\delta\mathbf{x}^*$, for minimizing the cost function, $J$, is the solution to
$\mathbf{A}\delta\mathbf{x}^* = \mathbf{b}$. Each element of the state is then updated according to
\begin{equation*}
\begin{aligned}
{^\mylabel{}\transform[{\mathrm{op},{C_k}{C_1}}]} &\leftarrow\exp(\boldsymbol{\epsilon}_{k}^{*^\wedge}){^\mylabel{}\transform[{\mathrm{op},{C_k}{C_1}}]}\\ 
\mathbf{p}_{\mathrm{op},C_1}^{{j_1}{C_1}} &\leftarrow \mathbf{p}_{\mathrm{op},C_1}^{{j_1}{C_1}} + \mathbf{D}\boldsymbol{\zeta}^{*}_{j},
\end{aligned}
\end{equation*}
where the vector operator $\left(\cdot\right)^{\wedge}$ is defined in \cite{barfoot2017}. The cost function is then relinearized about the updated operating point and the process iterates until convergence.

\subsection{Egocentric and Geocentric Trajectories} \label{sec:geocentric}
\paragraph{Egocentric}
Egocentric motions are expressed in the moving camera frame, $\protect\underrightarrow{\boldsymbol{\mathcal{F}}}_{C_k}$. The egocentric motion of the camera is identity by definition and the egocentric motions of the scene are given by
\begin{equation*}
\forall\mylabel{}\in\mathcal{L}, {^{\mathrm{ego}}\transform[{{\ell_K}{\ell_1}}]} \coloneqq {^\mylabel{}\transform[{{C_K}{C_1}}]^{-1}}.
\end{equation*}
One of these motions is the egocentric motion of the static world \emph{caused} by the camera motion.
\paragraph{Geocentric}
Geocentric motions are expressed in some earth-attached frame. The geocentric motion of the camera is given by the hypothesis motion estimated from the static background,
\begin{equation*}
T_{{C_K}{C_1}} \coloneqq \left.{^\mylabel{}T_{C}} \right|_{_{\mylabel{}=\mylabel{}_{\mathrm{static}}}},
\end{equation*}
where the static label may be selected by heuristics as in VO (e.g., label support size).

\begin{figure}[t]
	\centering
	\includegraphics[clip,width=0.98\columnwidth,page=1]{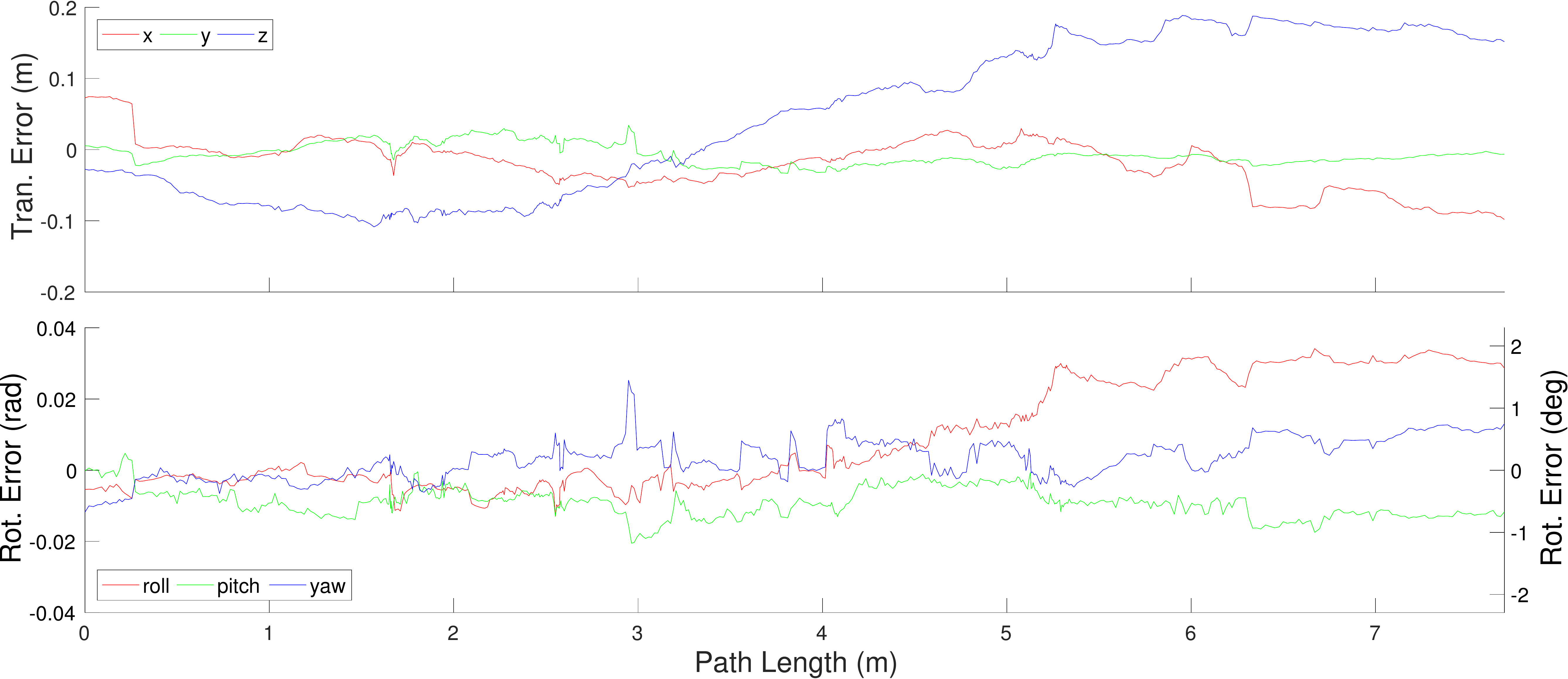}%
	\caption{The translational and rotational errors for the egomotion of the camera over its path compared to ground-truth trajectory data. Errors are reported in the egocentric camera frame with the z-axis along the optical axis, and the y-axis down.}
	\label{fig:egomotion}
\end{figure}

The geocentric motions of the rest of the scene are given by\looseness=-1
\begin{equation*}
\forall\mylabel{}\in\mathcal{L}\setminus\mylabel{}_{\mathrm{static}},\; \transform[{\mylabel{}_{k}\mylabel{}_{1}}] = \mathbf{F}_{{\mylabel_k}{\mylabel_1}} \transform[{\mylabel{}_{1}C_{1}}]{^\mylabel{}\transform[{{C_k}{C_1}}]^{-1}} \transform[{C_{k}C_{1}}] \transform[{\mylabel{}_{1}C_{1}}]^{-1},
\end{equation*}
where $\mathbf{F}_{{\mylabel_k}{\mylabel_1}}$ is the object deformation matrix and is assumed to be identity, (i.e., rigid body). The initial transform,
\begin{equation*}
\transform[{\mylabel{}_{1}C_{1}}] = \begin{bmatrix}
\mathbf{C}_{\mylabel_1 C_1} & \mathbf{r}_{\mylabel{}_{1}}^{C_{1}\mylabel{}_{1}}\\
\mathbf{0}^T & 1
\end{bmatrix},
\end{equation*}
relates the camera to the center of motion of each object, $\mathbf{r}_{\mylabel{}_{1}}^{C_{1}\mylabel{}_{1}}$. The object center is calculated from the centroid of all points, $\mathcal{P}_{\mylabel{}}$, projected into the first observed frame,
\begin{equation*}
\mathbf{r}_{\mylabel{}_{1}}^{C_{1}\mylabel{}_{1}} = -\frac{1}{\vert\mathcal{P}_\mylabel{}\vert}\mathbf{C}_{\mylabel_1 C_1}\sum_{j = 1}^{\vert\mathcal{P}_\mylabel{}\vert}{ {^{\mylabel{}}\transform[{{C_{t_j}}{C_1}}]^{-1}}\mathbf{p}_{C_{t_j}}^{{j_{t_j}}{C_{t_j}}}},
\end{equation*}
where $t_j$ is the first frame where $p^j$ is observed, and $\mathbf{C}_{\mylabel_1 C_1}$ is arbitrary and assumed to be identity. This averaging allows the centroid estimate to adjust as new points are observed due to rotation or occlusion.

\section{Experiments and Results} \label{sec:experiments}
The accuracy of the MVO algorithm is evaluated on real-world data collected using a Bumblebee XB3 stereo camera and a Vicon motion capture system. Unlike existing multimotion datasets designed for egomotion or segmentation (e.g., \cite{geiger2012, tron2007}), this new dataset contains ground truth for the \emph{entire} scene which consisted of a moving camera observing four moving blocks (\cref{fig:marquee}).

The results (\cref{fig:egomotion,fig:results}) were produced from a $500$-frame image sequence. Estimation was performed as a 48-frame sliding window, with $5$ neighbors for each point in the graph, $1000$ RANSAC iterations per new label, $e_{\mathrm{th}} = 4$, $\alpha=100$, $\beta=2$, $\psi_{\ell}=1000$, $\lambda = 1$, and a minimum model size and length of $10$ points and $3$ frames, respectively. Feature detection and matching were performed using LIBVISO2 \cite{geiger2011} and the Gauss-Newton minimization was performed with Ceres \cite{ceres-solver} using analytical derivatives (\cref{sec:batch}). 

The transforms between the Vicon frames and our estimated frames are arbitrary, so the first $25$ frames of the estimates are used to calibrate this transform. All errors are reported for geocentric trajectory estimates.
The camera egomotion (\cref{fig:egomotion}) exhibits a maximum total drift of $0.21$ m, $3.24\%$ of total path length, and a maximum rotational error of $1.96^{\circ}$, $-1.00^{\circ}$, and $-0.42^{\circ}$ in roll-pitch-yaw, respectively. This error is reasonable compared to the level of drift in other model-free, camera-only VO systems \cite{geiger2012}. \looseness=-1 

The motion estimates of the bodies varied with their motion and their visibility.
The two swinging blocks partially left the camera frustum near the beginning of the segment which caused estimation dropouts and higher translational errors.
A portion of the geocentric error of each motion is due to the error in the camera motion estimate. The maximum translational and rotational errors for each block are $0.44$ m, $-9.09^{\circ}$, $-5.28^{\circ}$, and $2.67^{\circ}$ for the top-left block; $0.27$ m, $-6.19^{\circ}$, $-6.05^{\circ}$, and $-5.00^{\circ}$ for the top-right block; $0.99$ m, $2.33^{\circ}$, $6.58^{\circ}$, and $3.15^{\circ}$ for the bottom-left block; and $0.39$ m, $-2.74^{\circ}$, $1.71^{\circ}$, and $-8.10^{\circ}$ for the bottom-right block (\cref{fig:results}). \looseness=-1

The gaps in the trajectories occur when the motion was not successfully segmented or the final estimation stage (\cref{sec:batch}) fails to converge. This is often due to poor feature distribution, especially when objects reach the edge~of~the camera frustum. These discontinuities, coupled with~the dynamic camera motion, caused errors in the trajectory~estimate. 

\section{Discussion} \label{sec:discussion}
MVO consistently segments and estimates the motions of the camera and the four independent objects when they are fully visible. As is to be expected, the two blocks that partially exited the camera frustum had segmentation failures and higher errors, largely due to incomplete feature tracklets. 

Feature distribution is an important factor in the performance of any sparse approach \cite{farboud2014}. This problem is exacerbated for dynamic objects that take up a small portion of the scene. Additionally, the appearance of these objects is often more volatile making feature association even more difficult. The algorithm is dependent on the accuracy of the input tracklet set and cannot estimate motions for which there are insufficient features. Feature dropouts and lack of tracklets are therefore significant challenges for this type of pipeline, and the development of more robust feature detection and matching pipelines is an ongoing area of research.

The distribution of features also influences the estimate of the center of the motion. It is difficult to infer the structure of an object beyond the surfaces that are observed in the batch without \textit{a priori} knowledge of the object's shape. This means the centroid of the observed feature points for a label can often be a bad estimate of the the label's true center of motion. \looseness=-1

\section{Conclusion} \label{sec:conclusion}
This paper extends the classic VO pipeline to address the multimotion estimation problem. 
The multimotion visual odometry (MVO) pipeline segments and estimates \emph{all} rigid motions in a scene. 
It does so by using feature-tracking, sparse graph segmentation, and multiframe batch motion estimation such that it avoids many of the limitations of other multimotion estimation approaches.

We evaluated MVO on a multimotion dataset with ground-truth trajectories for all motions in the scene. Its estimation accuracy is comparable to similarly defined egomotion-only VO systems while also exhibiting similar limitations. We are actively exploring the application benefits of continuous-time state estimation and continuous labels, as well as implementing the pipeline such that it can be used in real-time.\looseness=-1

\section*{Acknowledgment}
We would like to thank Paul Amayo for his insightful conversations on convex optimization and multilabeling, and for his implementation of CORAL.

\end{document}